%% file: abstract_template.tex
\documentclass{llncs}
\usepackage{color}
\usepackage{mathptmx}       
\usepackage{helvet}         
\usepackage{courier}        
\usepackage{type1cm}        
\usepackage{makeidx}         
\usepackage{graphicx}        
\usepackage{multicol}        
\usepackage[bottom]{footmisc}
\usepackage{subfigure}
\usepackage{amsfonts}
\usepackage[cmex10]{amsmath}
\usepackage{lipsum}
\newcommand*\samethanks[1][\value{footnote}]{\footnotemark[#1]}
\newcommand{\matr}[1]{\mathbf{#1}}
\usepackage{textpos} 

\begin{document}

\title{Graphlet Count Estimation via Convolutional Neural Networks}

\titlerunning{Hamiltonian Mechanics}  
%
\author{Xutong Liu\inst{1}\thanks{Both authors contributed equally to this work} \and Yu-Zhen Janice Chen\inst{1}\samethanks \and
	John C.S. Lui\inst{1} \and Konstantin Avrachenkov\inst{2}}

\authorrunning{Xutong Liu, Yu-Zhen Janice Chen et al.} 

\institute{
	The Chinese University of Hong Kong
	\and
	Inria Sophia Antipolis, France}

\maketitle    
\begin{textblock*}{3cm}(-0.6cm,16.5cm)
	\setlength{\fboxrule}{0pt}
	\fbox{\footnotesize Extended Abstract Accepted by COMPLEX NETWORKS 2018}
\end{textblock*}

\input{intro.tex}

\input{method.tex}

\input{data.tex}
\input{result.tex}

\end{document}

%% file: intro.tex
\vspace{-0.4cm}
\noindent
\paragraph{\bf 1 \hspace{0.05in} Introduction.} 
Graphlets are defined as $k$-node connected induced subgraph patterns. 
For an undirected graph, 3-node graphlets include close triangle (\includegraphics[height=\fontcharht\font`\B]{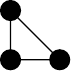}) and open triangle (\includegraphics[height=\fontcharht\font`\B]{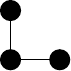}).
When $k = 4$, there are six different types of graphlets, e.g., tailed-triangle (\includegraphics[height=\fontcharht\font`\B]{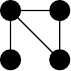}) and clique (\includegraphics[height=\fontcharht\font`\B]{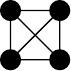}) are two possible $4$-node graphlets. 
The number of each graphlet, called graphlet count, is a signature which characterizes the {\it{local} network structure} of a given graph.
Graphlet count plays a prominent role in network analysis of many fields, most notably bioinformatics~\cite{Przu07} and  social science~\cite{ChLi16}.

However, enumerating exact graphlet count is inherently difficult and computational expensive because
the number of graphlets grows exponentially large as the graph size and/or graphlet size $k$ grow \cite{ChLi16}. 
To deal with this difficulty, many sampling methods were proposed for estimating graphlet count with bounded error~\cite{BrChKuLePa17,ChLi16,RaBhAl14}. 
Nevertheless, these methods require large number of samples to be statistically reliable, which is still computationally demanding. 
Moreover, they have to repeat laborious counting procedure 
even if a new graph is similar or exactly the same as previous studied graphs. 

Intuitively, learning from historic graphs can 
make estimation more accurate and avoid many repetitive counting to reduce computational cost. 
Based on this idea, we propose a {\em convolutional neural network} (CNN) framework and 	
two {\em preprocessing techniques} to estimate graphlet count.\footnote{Our code is accessible at https://github.com/jjanicechen/GraphletCountEstimationCNN.git}
Extensive experiments on two types of random graphs and real world biochemistry graphs 
show that our framework can offer substantial speedup on estimating graphlet count 
of new graphs with high accuracy.

%% file: method.tex
\vspace{-0.09cm}

\noindent
\paragraph{\bf 2 \hspace{0.05in} Method.} 
Given a set of undirected graphs and a particular type of $k$-node graphlet, 
our objective is to develop a CNN which will be trained using part of dataset with known graphlet counts. 
After training, the CNN can quickly and accurately predict graphlet counts of other unseen graph samples in the set. 
Our framework takes the graph adjacency matrix as input 
and outputs the graphlet count of the input graph.
Let us define some notations for our CNN.
Let $\matr{O}^{(l)} \in \mathbb{R} ^ { N^{(l)} \times N^{(l)} \times C^{(l)}}$ be the output tensor at layer $l$, where $l=0,1,2,3$, $N^{(l)}$ denotes the width (and height) along each channel and $C^{(l)}$ denotes the channel size. Let $\matr{O}_{i,j,t}^{(l)}$ be the $(i,j)^{\textnormal{th}}$ element along the $t^{\textnormal{th}}$ channel.
We assign $\matr{O}^{(0)}$ as the graph adjacency matrix.
Mathematically, our CNN structure can be described as follows:

\vspace{-0.22cm}
\resizebox{0.95\columnwidth}{!}{
\begin{minipage}{\columnwidth}
\begin{align}
\matr{O}_{i,j,t}^{(l)} &= {\textnormal{ReLU}}(\matr{W}_t^{(l)}\cdot \matr{O}^{(l-1)}[i: i+H^{(l)}-1,\, j : j+H^{(l)}-1,\: :\:] + b_t^{(l)}),\quad l = 1,2, \label{eq1} \\
\matr{O}^{(3)} &=  {\textnormal{ReLU}}({\textnormal{Flatten}}(\matr{O}^{(2)})^T \matr{W}^{(3)} + b^{(3)}). \label{eq2}
\end{align} 
\end{minipage}
}

Equation (\ref{eq1}) corresponds to two convolution layers. Each layer applies $C^{(l+1)}$ filters over the input feature map $\matr{O}^{(l-1)}$, and the $t^{th}$ filter is parameterized by a trainable 
3D weight tensor $\matr{W}_t^{(l)} \in \mathbb{R} ^ {H^{(l)} \times H^{(l)} \times C^{(l)}}$, where $H^{(l)}$ denotes the width (and height) of the filter. $[a:b, c:d, :]$ is a slicing function which extracts subset of elements indexing from $a$ to $b$ in width, $c$ to $d$ in height and all in channel to form a new tensor. $\cdot$ is the sum of element wise product of two tensors. 
After adding bias term $b_t^{(l)}$, we apply ReLU ($\textnormal{max}(0,x)$) as the activation function to obtain the output feature map 
$\matr{O}^{(l)}$.
Equation (\ref{eq2}) is associated with the fully connected layer. 
It flattens the output $\matr{O}^{(2)}$ 
into a column vector, applies $\matr{W}^{(3)}$, $b^{(3)}$ and ReLU to obtain the estimated graphlet count. 
Finally, our CNN is trained with back propagation and mean squared error as the loss function.

The above CNN structure inherits the learning power for local structural information of graphs. 
However, we still need to address the following challenges: 
(1) The input adjacency matrix is not consistent because graphs in the training set may have different sizes.
(2)  In practice, real world network dataset may not contain sufficient amount of graph samples for training, which will cause overfitting problem.
To address these challenges, we introduce two preprocessing techniques:\\
\noindent {\textbf{\textit{Adjacency Matrix Zero Padding.}}} 
To preserve edge connectivity information of all training graphs, we consider the largest graph in the training set,
and use its dimension (say $N$) as the dimension of the input adjacency matrix ($N \times N$).
For other graphs in the training set, we take each adjacency matrix and pad it with zero till we have
an input matrix of dimension $N \times N$. This solves the varying input size problem.\\
\noindent \textbf{\textit{Swapping Augmentation.}}
To acquire sufficient data for training, we take advantage of the graph isomorphism property, where a graph can be expressed by different input adjacency matrices having the same underlying network structure.
Our approach is to randomly pick indices $i$ and $j$, then swap the $i^{\textnormal{th}}$ row with $j^{\textnormal{th}}$ row and $i^{\textnormal{th}}$ column with $j^{\textnormal{th}}$ column of the adjacency matrix. 
We can repeat the swapping operation for each graph $m$ times to create $m$ more training data. 
Analogous to flipping or rotation of images, we improve CNN's generalization ability and thus improve the accuracy of our model.

%% file: data.tex
\noindent
\paragraph{\bf 3 \hspace{0.05in} Data and Metric.} 
Here, we introduce our testing datasets,  benchmarking works, and evaluation metrics.

\noindent\textbf{\textit{Random Graph.}} 
We synthesize datasets with two random graph models: random geometric graph (RGG) and Erdos-Renyi (ER) graph. 
A RGG is constructed by placing nodes uniformly at random in a unit cube and connecting two nodes by an edge if and only if their distance is within a given radius $r$. 
In a ER graph, the edge between every two nodes exists with probability $p$.
In each synthetic dataset, we have 3000 training graphs, 300 validation graphs, and 300 testing graphs.

\noindent\textbf{\textit{Empirical Network.}} 
We test on three real world biochemistry datasets: MUTAG \cite{ViScKoBo10}, NCI1 and NCI109 \cite{WaWaKa08}. 
MUTAG dataset contains 188 mutagenic compound graphs. 
NCI1 and NCI109 each has 4110 and 4127 chemical compound graphs tested on lung and ovarian cancer cells respectively. 
For MUTAG, we use \textit{swapping augmentation} to increase the number of training samples. 
We also apply \textit{adjacency matrix zero padding} to make all graphs in each dataset have the same size.

\noindent\textbf{\textit{Benchmark.}} 
We compare CNN with three existing frameworks: GRAFT \cite{RaBhAl14}, CC2 \cite{BrChKuLePa17}, GUISE \cite{BhRaAl}, which are based on edge sampling, color coding, and Markov Chain Monte Carlo method respectively.

\noindent\textbf{\textit{Relative Error.}}
Let $c_i$ be the ground truth graphlet count of sample graph $i$, $c'_i$ be its estimated count, and there are $S$ samples in the dataset.
We compute the mean absolute error of the estimations, ${\textnormal{mae}} = \Sigma_{i=1}^{S} \vert c'_{i} - c_{i}\vert / S$, and mean of ground truth counts, $\mu = \Sigma_{i=1}^{S} c_i / S $. We take relative error as $e = {\textnormal{mae}}  /\mu$ .

\noindent\textbf{\textit{Speed.}}
To ensure a fair comparison, we do not choose running time as the performance metric since it highly depends on hardware and implementation (e.g. running on GPU/CPU). 
Instead, we measure the number of arithmetic operations they use. 
For CNN model, we compute the number of floating-point operations (FLOPs). 
For benchmarking works, we calculate the number of comparison operations in the algorithms.

%% file: result.tex
\noindent
\paragraph{\bf 4 \hspace{0.05in} Result.} 
We test our framework on random graph datasets. 
For approximating 4-clique counts, our CNN model achieves less than 8\% relative error on 50-node RGGs with radius 0.45 and less than 5\% relative error on 50-node ER graphs with edge existing probability 0.5. 
We also train our CNN models for estimating 4-path (\includegraphics[height=\fontcharht\font`\B]{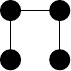}), 3-star (\includegraphics[height=\fontcharht\font`\B]{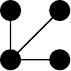}), 5-path (\includegraphics[height=\fontcharht\font`\B]{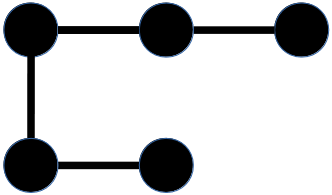}) on the empirical biochemistry datasets. 
The relative errors on all three datasets are less than 20\% of the ground truth counts. 
For estimating 4-path on MUTAG dataset, our model performs especially well making only 6\% relative error.  

\begin{figure}
	\centering
	\vspace{-0.6cm}
	\includegraphics[scale=0.26]{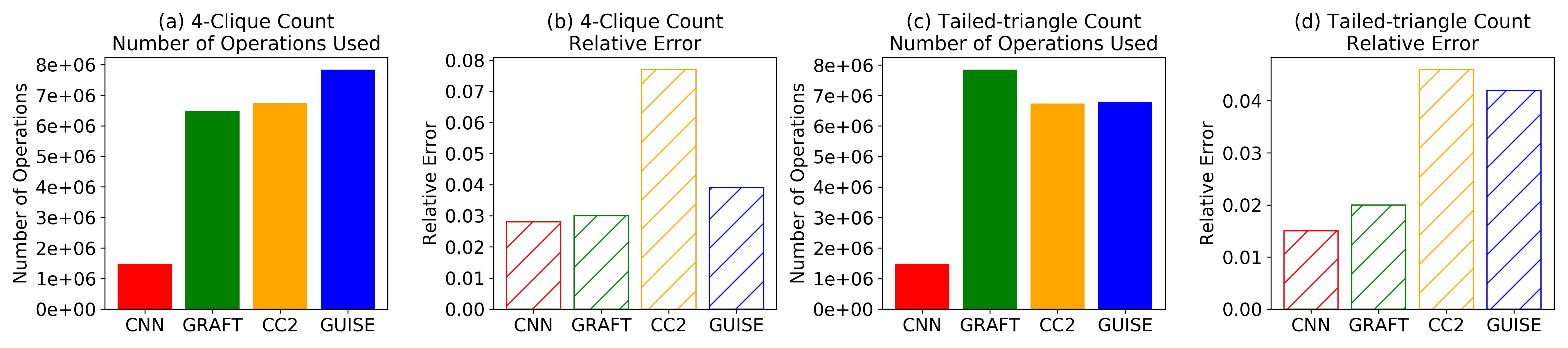}
	\vspace{-0.35cm}
	\caption{Comparison of the number of arithmetic operations used for estimating 4-clique counts, tailed-triangle counts on 50-node ER graphs with edge existing probability 0.5. (a, c) The number of operations used by each framework. (b, d) The relative error each framework makes.}
	\label{fig:comp}       
	\vspace{-0.68cm}
\end{figure}

To compare the speed of our CNN with existing methods, the number of arithmetic operations used are calculated. 
For a fair comparison, we tune the number of iterations for all benchmarking sampling methods, so that they obtain as close relative errors to that of CNN as possible.
Figure~\ref{fig:comp} (a, c) shows that the numbers of arithmetic operations used by GRAFT, CC2, or GUISE are significantly more than that used by CNN. 
This result demonstrates that our CNN based graphlet count estimation approach offers remarkable speedup on predicting graphlet counts while still maintaining high accuracy.